\documentclass[conference]{IEEEtran}
\IEEEoverridecommandlockouts
\usepackage{cite}
\usepackage{amsmath,amssymb,amsfonts}
\usepackage{algorithmic}
\usepackage{graphicx}
\usepackage{textcomp}
\usepackage{xcolor}
\def\BibTeX{{\rm B\kern-.05em{\sc i\kern-.025em b}\kern-.08em
    T\kern-.1667em\lower.7ex\hbox{E}\kern-.125emX}}
\begin{document}

\title{Image Copy-Move Forgery Detection via Deep Cross-Scale PatchMatch\\
\thanks{}}
%
\author{\IEEEauthorblockN{Yingjie He, Yuanman Li*, Changsheng Chen and  Xia Li}
\IEEEauthorblockA{Guangdong Key Laboratory of Intelligent Information Processing, College of Electronics and Information Engineering,\\ Shenzhen University, China}
\IEEEauthorblockA{2110436087@email.szu.edu.cn, yuanmanli@szu.edu.cn, cschen@szu.edu.cn, lixia@szu.edu.cn                }
}
%

\maketitle

\begin{abstract}
The recently developed deep algorithms achieve promising progress in the field of image copy-move forgery detection (CMFD). However, they have limited generalizability in some practical scenarios, where the copy-move objects may not appear in the training images or cloned regions are from the background. To address the above issues, in this work, we propose a novel end-to-end CMFD framework by integrating merits from both conventional and deep methods. Specifically, we design a deep cross-scale patchmatch method tailored for CMFD to localize copy-move regions. In contrast to existing deep models, our scheme aims to seek explicit and reliable point-to-point matching between source and target regions using features extracted from high-resolution scales. Further, we develop a manipulation region location branch for source/target separation. The proposed CMFD framework is completely differentiable and can be trained in an end-to-end manner. Extensive experimental results demonstrate the high generalizability of our method to different copy-move contents, and the proposed scheme achieves significantly better performance than existing approaches. 
\end{abstract}
\section{Introduction}
\label{sec:intro}
The rapid development of digital image editing tools makes image forgeries very common in our lives. Such forged images could be maliciously used in internet rumors, insurance fraud, fake news, and dishonest academic literature, causing serious security issues for society.   Copy-move forgery is one of the most common manipulations, which tries to add or hide objects of interest in an image, by copying one or several regions from an image and then pasting them elsewhere in the same image. Detecting image copy-move forgery sometimes is very challenging because the forged regions come from the same image, i.e., they share very similar statistical behaviors with the other untouched regions, such as the noise distribution, brightness, and photometric characteristic. 

\begin{figure*}[t!]
	\centering
	\includegraphics[width=0.9\linewidth]{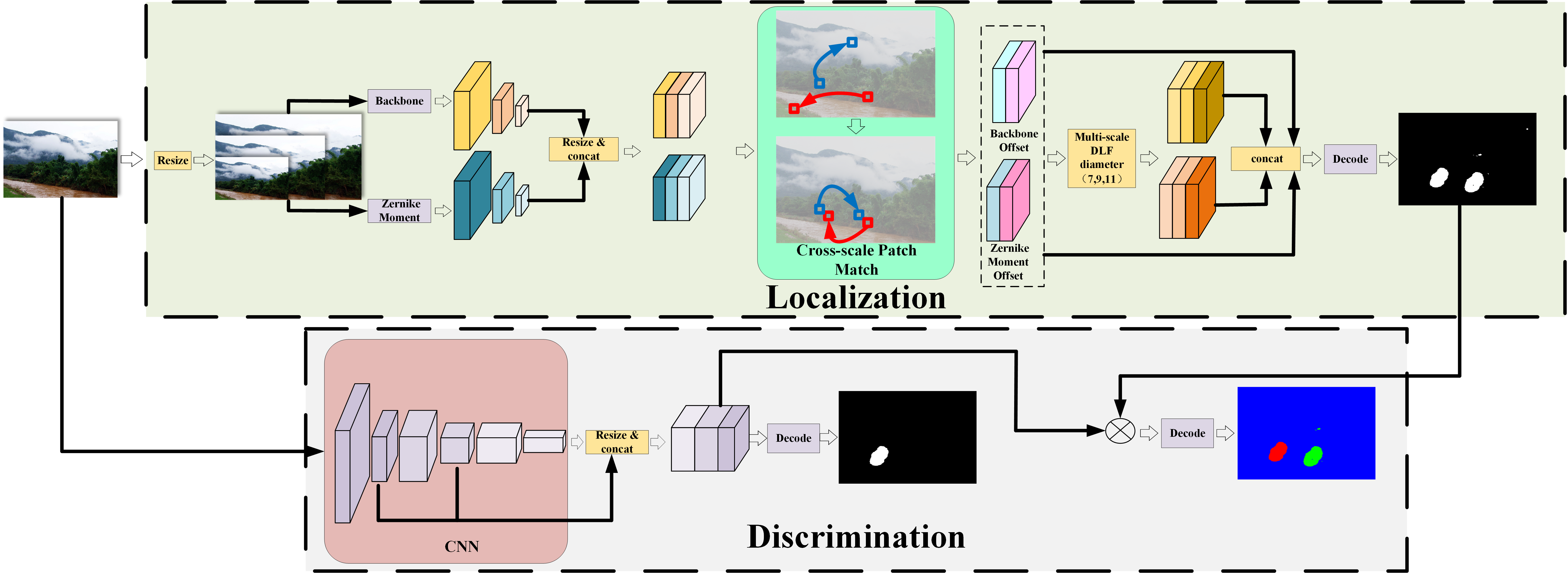}
	\centering
	\caption{The framework of our proposed algorithm. The top branch is used to localize copy-move regions via dense-field matching, and the bottom branch is used to differentiate the source and target regions.}
	\label{fig:framework}
\end{figure*}

As an important topic in the community of multimedia security, image copy-move forgery detection (CMFD) has been widely studied in the past years, and a number of approaches have been proposed from different aspects. Conventional CMFD \cite{cozzolino_efficient_2015,ryu_rotation_2013,li_fast_2019} approaches employ hand-crafted features to identify the copy-move correspondences, such as the block-based methods and keypoint-based methods. The forged traces revealed by conventional algorithms are often trustworthy since copy-moved correspondences can be explicitly linked through block matching or keypoint matching.
However, these methods heavily rely on the selected hand-crafted featured designed for common computed vision tasks. As a consequence, the conventional methods are shown to be less effective in detecting complex copy-move forgeries and are also fragile against to post-processing. Further, all the existing conventional methods can detect only the copy-move correspondences, while failing to discriminate source/target regions. 

Motivated by the powerful representation capability of deep features, some deep CMFD frameworks have been studied and attracted increasing attention in recent years\cite{li2022image}. Wu \textit{et al.} \cite{ferrari_busternet_2018} proposed the first end-to-end CMFD framework with source/target separation. Different from conventional algorithms, features of  \cite{ferrari_busternet_2018} are adaptively learned by the convolutional neural network (CNN) over a CMFD dataset rather than manually designed. Inspired by \cite{ferrari_busternet_2018}, many follow-up deep CMFD models have been proposed, such as \cite{Yuan2016WIFS,chen_serial_2020,islam_doa-gan_2020,zhong2022SC}. Equipped with representative features, these deep CMFD methods have been shown to be more effective in reality and more robust against post-processing, such as additive gaussian noise and compression.
Despite their advantages, there still exist some fundamental limitations of existing deep CMFD approaches as below:
\begin{itemize}
    \item Most existing deep CMFD approaches reveal the suspicious copy-move regions by calculating an attention map using high-level features learned by deep CNNs. Such features of low resolutions are often overfitted to objects in the training images, thus severely reducing the generalizability of the resulting methods. As a result, existing approaches are less effective when the copy-moved objects do not appear in the training set. Even worse, they may completely fail when the copy-move process was conducted in the background or using incomplete objects.
    \item Unlike the conventional algorithms, the deep CMFD approaches fail to establish explicit point-to-point matching between copy-move correspondences. This makes their detected results lack interpretability and reliability. 
\end{itemize}

To address the above issues, in this work, we propose a novel end-to-end CMFD framework based on Deep PatchMatch (DPM), which integrates merits from both conventional methods and deep models. On the one hand, DPM aims to model the point-to-point matching between source and target regions in high-resolution scales, which makes our method highly generalizable for different types of copy-move regions. On the other hand, DPM is completely differentiable, then the features can be adaptively learned in an end-to-end training process.

Our main contributions are summarized as follows:
\begin{itemize}
\item We propose a novel end-to-end deep CMFD framework with source/target separation, which has advantages of both conventional and existing deep CMFD models. Our framework is of high generalizability to the copy-move contents, including objects, incomplete objects, and even background. 

\item We develop a new similarity localization method based on differentiable patchmatch tailored for CMFD, where both conventional features and CNN features are adopted. Several schemes, such as cross-scale matching scheme and multi-scale dense fitting error estimation are devised for seeking reliable point-to-point matching between source and target regions in high-resolution scales. We further design a sample branch for source/target separation.

\item We provide a series of experiments on two widely used challenging benchmark datasets, and the result results show that our model outperforms recent state-of-the-art methods.
\end{itemize}

The remainder of this paper is organized as follows. We detail our proposed CMFD framework in Section \ref{sec:proposed}. Extensive experimental results and ablation studies are given in Section \ref{sec:Experiment Result}, and Section \ref{sec:Conculsion} draws the conclusion.

\section{Proposed Method}
\label{sec:proposed}
\subsection{Dense-Field Matching via Deep Cross-Scale PatchMatch (DFM)}

As illustrated in the top of Fig. \ref{fig:framework}, DFM detects copy-move regions through point-to-point matching using features maps of high-resolutions which prevents our model from overfitting to objects. Besides, our model can effectively integrate the advantages of both conventional and deep models. The DFM branch consists of two blocks, i.e., 1) Feature extraction, 2) Cross-scale matching and prediction.

\subsubsection{Feature Extraction}
The feature extraction consists of a pyramid feature learning block and a Zernike Moments\cite{1980Image} extraction block.  To fight against severe rescaling attacks,  we propose a pyramid feature extraction. Specifically, given an image ${I}_{o}\in{R}^{H\times{W}\times{3}}$, we first downsample  or upsample it by
\begin{equation}
    I_b = resize(I_o, r_b), ~~I_u=resize(I_o,r_u),
\end{equation} where $r_b$ and $r_u$ are set to $0.75$ and $1.5$. 
The architecture of our backbone includes five convolution blocks, and each block consists of a convolution, a BatchNorm, and a ReLu layer. The backbone has a resizing layer in the end to rescale features maps to the same dimension $H\times W\times c$.  Finally, we obtain three features maps as  
\begin{equation}
    F_u = \mathcal{F}_{et}(I_u), ~F_o = \mathcal{F}_{et}(I_o), ~F_b = \mathcal{F}_{et}(I_b),
\end{equation} where $\mathcal{F}_{et}(\cdot)$ denotes the backbone feature extraction function.  

In our experiment, we find that a pure CNN feature extraction from random initialization is very difficult to learn rotation-invariant features. In this work, we also adopt Zernike Moments as a complement to CNN features, which have desirable robustness against rotations. We obtain three Zernike Moments features maps as
\begin{equation}
    F'_u = \mathcal{F}_{zm}(I_u), ~F'_o = \mathcal{F}_{zm}(I_o), ~F'_b = \mathcal{F}_{zm}(I_b).
\end{equation}
It also has a resizing layer in the end to rescale features maps to the same dimension $H\times W\times 12$.

\subsubsection{Deep Cross-Scale PatchMatch and prediction} 
Finding reliable matches among pixels is crucial in copy-move forgery detection. Given an image $I \in R^{W\times H\times 3}$, we define $\delta: (i,j) \rightarrow R^2$ as an offset function determined by the nearest-neighbor field over all possible pixel coordinates. Formally, 
\begin{equation}\label{eq:offset}
	\begin{aligned}
	    \delta(i,j) =\arg\min_{i_s,j_s\neq 0} Dis(f(i,j), f(i+i_s,j+j_s)),
	\end{aligned}
\end{equation} where $0 \leq i+i_s\leq M$, $0 \leq j+j_s\leq N$, $f(i,j)$ denotes the feature vector for the location $(i,j)$, and $Dis(\cdot)$ is an appropriate distance measurement. Apparently, finding the dense matches among pixels is equivalent to solving the optimization problem (\ref{eq:offset}). For simplicity, we write $\delta(i,j) = (\delta_{ij}^x, \delta_{ij}^y)$. Then, the relationship between $(i,j)$ and $\delta(i,j)$ can be written as
\begin{equation}
		\begin{aligned}
		 (i,j) = match\left(i+\delta_{ij}^x,j+\delta_{ij}^y\right).
		\end{aligned}
\end{equation}

The PM \cite{barnes2009ToG} is an efficient algorithm to give an approximate solution to the problem (\ref{eq:offset}), by using the spatial regularity of natural images. Despite its efficiency, the classic PM method cannot be directly integrated into deep learning frameworks due to the non-differentiability of its design. Motivated by \cite{Shivam2019ICCV}, in this work, we design a differentiable deep cross-scale PM module to compute reliable point-to-point matching between source and target regions.
Compared with \cite{Shivam2019ICCV}, we re-design several procedures of PM to make it suitable for CMFD task. 
Cross-scale PM consists of the following three main layers.

\textbf{i. Initialization layer:} At the first step, we randomly generate an offset $\delta(i,j)$ for each pixel $I(i,j)$, ensuring that $(i+\delta_{ij}^x,j+\delta_{ij}^y)$ is still a valid location in the image. We discard $\delta(i,j)=(0,0)$ to remove trivial solutions. Note that with a large number of random offsets, it is very likely that some of them are optimal or near-optimal. 

\textbf{ii. Propagation layer:} This layer aims to propagate offsets of pixels to their neighbors. Our propagation layer consists of a propagation block and a random search block.
\begin{figure}	
	\centering
	\begin{minipage}[t]{0.5\linewidth}
        \centering
        \includegraphics[width=\textwidth]{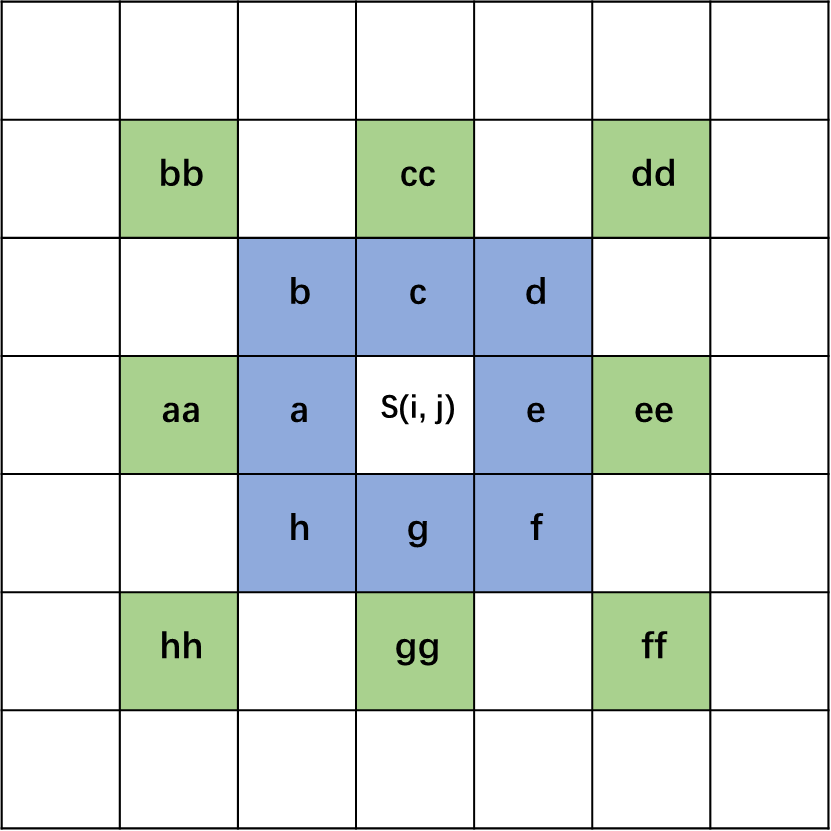}
    \end{minipage}
\caption{The colorful pixels propagate their offsets to the pixel (i,j).}
\label{fig:predictors}
\end{figure}

First, we consider the zero-order propagation along image rows and columns. As shown in Fig. \ref{fig:predictors}, we generate four zero-order candidate offsets for each pixel as follows
\begin{equation}\label{eq:zero-order}
		\begin{aligned}
		\delta^{\gamma}(i,j) &= \delta(i^{\gamma},j^{\gamma}),\\
		\gamma &\in \{a, c, e, g\},
		\end{aligned}
\end{equation}
where ($i^{\gamma},j^{\gamma}$) denotes pixel coordinates shown in Fig.\ref{fig:predictors}. Note that Eq. (\ref{eq:zero-order}) can be efficiently implemented using the circular shift operation in a deep framework.

Zero-order propagation is effective in addressing CMFD with rigid translations; however, it is less effective for rotation and rescaling transformations. In this paper, we also consider the first-order propagation. Specifically,  eight first-order candidate offsets are generated for each pixel as below
\begin{equation}
		\begin{aligned}
		\delta^{\gamma\gamma}(i,j) &= 2\delta(i^{\gamma},j^{\gamma})-\delta(i^{\gamma\gamma},j^{\gamma\gamma}),\\
		\gamma &\in \{a, b, c, d, e, f, g, h\}.
		\end{aligned}
\end{equation}

To avoid getting stuck in the local optimum, we further apply a random search for each pixel. specifically, we randomly generate four more candidate offsets for each pixel denoted by $\delta^{r1}(i,j), \delta^{r2}(i,j), \delta^{r3}(i,j)$ and $\delta^{r4}(i,j)$
\begin{equation}
    \begin{aligned}
    \delta^{\gamma}(i,j) &= \delta(i,j) + \Delta \delta(\gamma),\\
    \gamma &\in \{r1, r2, r3, r4\}.
    \end{aligned}
\end{equation} where $\Delta \delta(\cdot)$ is randomly sampled within the pre-defined search space centered at $(i,j)$. The radius of the search region is empirically set to  25 in our work. 

With the above propagation procedures, we finally produce seventeen candidate offsets for each pixel 
\begin{equation}
    \boldsymbol{\delta} = \{\delta, \delta^{a},..., \delta^{g}, \delta^{aa},...,\delta^{hh},\delta^{r1},..., \delta^{r4}\}.
\end{equation} For simplicity, we write $\boldsymbol{\delta} = \{\delta^1, \delta^2,...,\delta^K\}$, where $K=17$.

\begin{table*}
    \centering
	\caption{Discrimination performance comparisons on Synthetic dataset} 
	\label{table:Synthetic} 
 \scalebox{1}{
\begin{tabular}{c|ccc|ccc|ccc|ccc}
\hline
Methods & \multicolumn{3}{c|}{Background} & \multicolumn{3}{c|}{source} & \multicolumn{3}{c|}{target}& \multicolumn{3}{c}{average} \\ \cline{2-13} 
                         & precision    & recall    & F1   & precision   & recall   & F1    & precision  & recall  & F1 & precision  & recall  & F1 \\ \hline
BusterNet\cite{ferrari_busternet_2018} & 0.9196 & 0.9934 & 0.9530 & 0.1017 & 0.0687 & 0.0685 & 0.2483 & 0.0991 & 0.1227&0.4232&0.3870&0.3814\\
BusterNet$^{*}$ & 0.9412 & 0.9965 & 0.9661 & 0.0231 & 0.0231 & 0.0054 & 0.8666 & 0.8230 & 0.8295&0.6103&0.6142&0.6003\\
DOA-GAN\cite{islam_doa-gan_2020} & 0.9229 & \textbf{0.9969} & 0.9573 & 0.0535 &0.0185 & 0.0218 & 0.2685 & 0.1152 & 0.1364&0.4149&0.3768&0.3718\\
DOA-GAN$^{*}$ & 0.9736 & 0.9958 & 0.9845 & 0.6463 & 0.3450 & 0.4083 & \textbf{0.9083} & \textbf{0.9197} & \textbf{0.9081}&0.8427&0.7535&0.7669\\
Serial Network\cite{chen_serial_2020} & 0.9212 & 0.9693 & 0.9412 & 0.1725 & 0.0913 & 0.1046 & 0.2064 & 0.1609 & 0.1549&0.4334&0.4071&0.4002\\
Serial Network$^{*}$ & 0.9408 & 0.9531 & 0.9450 & 0.1299 & 0.0688 & 0.0776 & 0.5980 & 0.5298 & 0.4962&0.5562&0.5172&0.5062\\
ours & \textbf{0.9886} &  0.9921 & \textbf{0.9902} & \textbf{0.7556} & \textbf{0.7263} & \textbf{0.7159} & 0.8959 & 0.8948 & 0.8854& \textbf{0.8800}& \textbf{0.8710}& \textbf{0.8638}\\
\hline
\end{tabular}}
\end{table*}

\textbf{iii. Evaluation layer} We design the evaluation layer to calculate the best offset for each pixel from the $K$ obtained candidates in the propagation layer.  In this work, we adopt $\ell_1$-norm to measure the difference between two feature vectors. Then, for each given offset $\delta^{k}(i,j)$, the matching score is computed as
\begin{equation}
	\begin{aligned}
		S^{k}(i,j) = -\left|\left|f(i,j)- f\left((i,j)+ \delta^{k}(i,j))\right)\right|\right|_1.
	\end{aligned}
\end{equation}
It should be emphasized that $\delta^{k}(i,j)$ is from a continuous space (see Eq. \ref{eq:hat_S}). In this work, we adopt bilinear interpolation to obtain the corresponding feature vector $f((i,j)+ \delta^{k}(i,j))$.
In this work, we adopt a relaxed version of the $argmax$ function to obtain the best offset with the largest matching score $S^{k}(i,j)$.
\begin{equation}\label{eq:hat_S}
	\begin{aligned}
		\delta^{'}(i,j) = \sum_{k=1}^K\delta^{k}(i,j)\times{\hat{S}^{k}(i,j)},
	\end{aligned}
\end{equation}where 
\begin{equation}
    \hat{S}^{k}(i,j) = \frac{\exp(\beta S^{k}(i,j))}{\sum_{k=1}^{K}\exp(\beta  S^{k}(i,j))},
\end{equation} where $\beta$ is a temperature parameter. 
Then, $\delta^{'}(i,j)$ is fed again into the propagation layer and is treated as the current offset for the next iteration.

We propose a cross-scale PM algorithm. Different from the standard method, we evaluate offsets across features of different scales. Specifically, for the CNN features $F_u, F_o$ and $F_b$, we compute the matching scores as
\begin{equation}
	\begin{aligned}
		S_{nm}^{k}(i,j) = -&\left|\left|F_n(i,j)- F_m\left((i,j)+ \delta^{k}(i,j)\right)\right|\right|_1.\\
		&~~~~~~~~~~~~~~~~~~n,m\in\{u, o, b\}.
	\end{aligned}
\end{equation}
We keep only the best cross-scale matching score for each $k$ to update the offset, i.e., 
\begin{equation}
	S^{k}(i,j) = \max(S_{nm}^{k}(i,j)), ~n,m\in\{u, o, b\}.
\end{equation} Note that the propagation layer and the evaluation layer will be recurrently executed several times to achieve a good offset. 
For simplicity, we denote the resulting offset obtained by CNN features as $\delta_1$, while the offset obtained by ZM features as $\delta_2$. 

\begin{table}
	\centering
	\caption{Localization performance comparisons on CASIA CMFD dataset} 
	\label{table:casia}  
 \scalebox{1}{
	\begin{tabular}{c|ccc}
		\hline
		Methods & Precision & Recall & F1 \\ \hline
		OverSeg\cite{pun_image_2015} & 0.4297 & 0.3666 & 0.3956\\
		HFPM\cite{li_fast_2019} & 0.5781 & 0.6585 & 0.6157\\
		PM\cite{cozzolino_efficient_2015} & 0.4734 & 0.4952 & 0.4842\\
		BustNet\cite{ferrari_busternet_2018} & 0.5571 & 0.4383 & 0.4556\\
		DOA-GAN\cite{islam_doa-gan_2020} & 0.5470 & 0.3967 & 0.4144\\
		DFIC\cite{zhong_end--end_2020} & 0.7089 & 0.5885 & 0.6429\\
        Serial Network\cite{chen_serial_2020} & 0.5308 & 0.4979 & 0.4768\\
		Ours & \textbf{0.8363} & \textbf{0.7385} & \textbf{0.7542}\\
		\hline
	\end{tabular}}
\end{table}

Ideally, the offset should exhibit smooth linear behavior in copy-move regions, while chaotic in other regions. Inspired by \cite{cozzolino_efficient_2015}, we evaluate the matching quality based on dense linear fitting (DLF). Specifically, for each pixel, We calculate the DLF with three different radiuses $\rho \in  \{7, 9, 11\}$, which yields three different fitting errors $\epsilon_1^2,\epsilon_2^2$ and $\epsilon_3^2$. Finally, the fitting errors and offsets are fed into a simple decoder and predict a single-channel copy-move mask $M\in{R}^{H\times{W}\times 1}$, which calculates binary cross-entropy loss with ground truth
\begin{equation}
    M = Decoder\left(\epsilon_1^2,\epsilon_2^2,\epsilon_3^2, \delta_1,\delta_2\right).
\end{equation} Our decoder consists of five convolution blocks with a BatchNorm and a ReLu activation, while the last convolution block is followed by a sigmoid function.

\subsection{Source/Target Discrimination via manipulated regions detection (STD)}
For the STD branch, we apply a backbone network to extract the deep feature of the input image. Our backbone network consists of tree blocks, each of which has two convolution layers followed by a maxpool layer. We resize and concatenate features of each block to form a feature map $F_d \in R^{h\times w\times c}$, where $h=H/8$, $w=W/8$ and $c=448$. The feature $F_d$ is fed into a simple decoder and predicts a single-channel manipulated mask $M_d\in{R}^{H\times{W}\times 1}$. We adopt the binary cross-entropy loss to supervise the learning of $M_d$. In this way, the feature $F_d$ will concentrate on forgery areas. Then, we use the mask $M$ to reduce the effect of those background pixels
\begin{equation}
    F_d^{'} = F_d \odot M,
\end{equation}
where $\odot$ is the element-wise product operation. The feature $F_d^{'}$ is sent to a decoder consisting of five convolution layers to predict a three-channel mask $M_d\in{R}^{H\times{W}\times 3}$, which calculates categorical cross-entropy loss with ground truth.

\begin{figure*}[t!]
	\centering
	\includegraphics[width=0.9\linewidth]{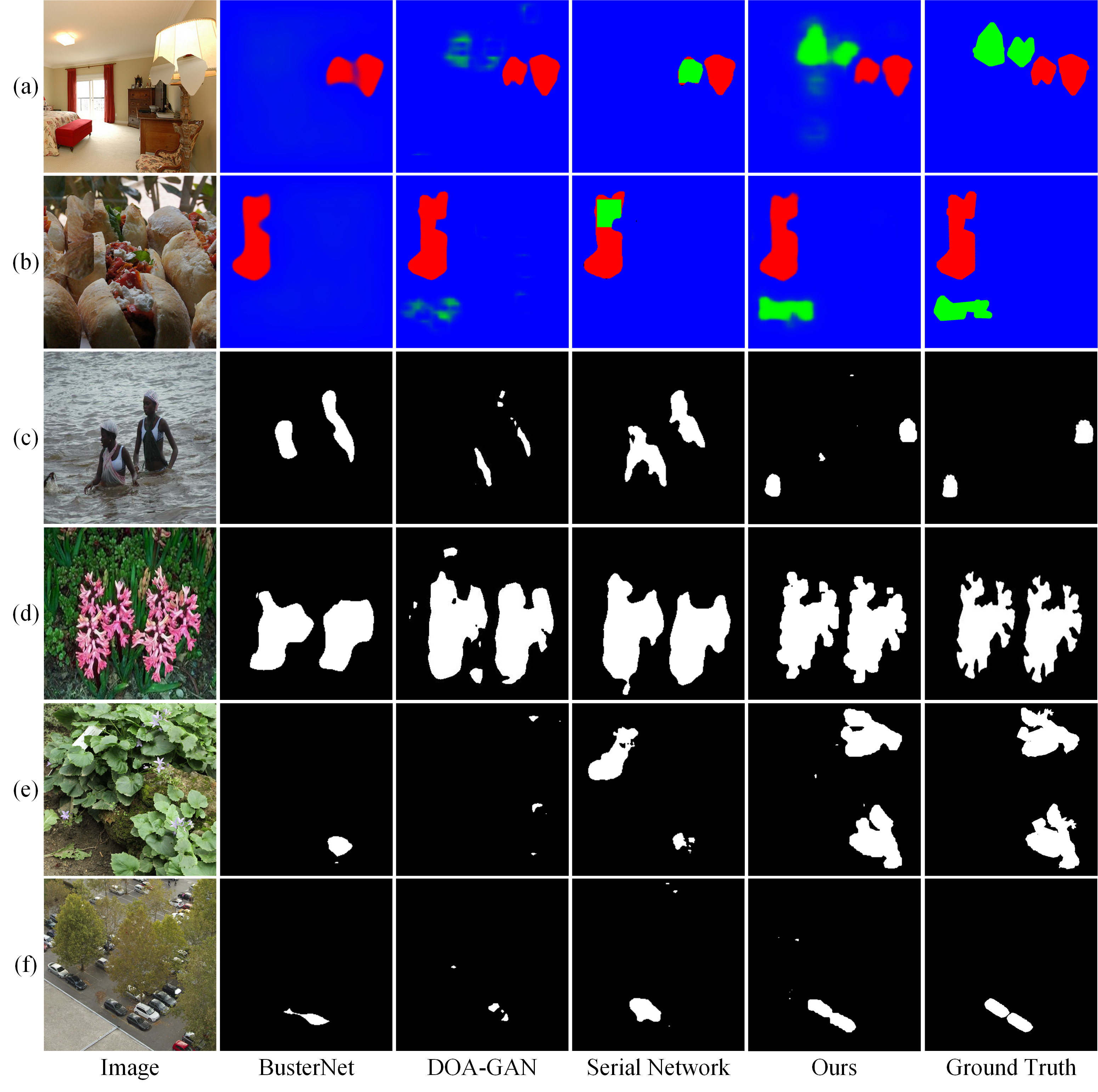}
	\centering
	\caption{Visualization examples from three datasets, in which the first and second images come from the Synthetic dataset, the 
third and fourth images come from CASIA CMFD dataset and the fifth and sixth come from CoMoFoD dataset.}
	\label{fig:visua}
\end{figure*}

\section{Experiments}\label{sec:Experiment Result}
\subsection{Implementation Details}
To train our model, we use the images from MS COCO 2014\cite{lin2014microsoft} to generate a synthetic training set. For each image, we resize it to 1024$\times$1024, select one or several random polygon regions and paste them into the rest of the image.  All copied snippets randomly suffered from the rotation and scaling attack. The attack parameters are: rotation $\in$ [-180°,180°], scale$\in$[0.5,2]. Following this rule, we generate 79,353 training images. For testing, 4,000 testing images are generated in the same way.

\begin{table}
	\centering
	\caption{Performance comparisons on CoMoFoD dataset} 
	\label{table:comofod}  
 \scalebox{1}{
	\begin{tabular}{c|ccc}
		\hline
		Methods & Precision & Recall & F1 \\ \hline
		OverSeg\cite{pun_image_2015} & 0.3491 & 0.2141 & 0.2220\\
		HFPM\cite{li_fast_2019} & 0.4280 & 0.4240 & 0.4260\\
		PM\cite{cozzolino_efficient_2015} & 0.4761 & 0.3992 & 0.4183\\
		BustNet\cite{ferrari_busternet_2018} & 0.5125 & 0.4167 & 0.4378\\
		DOA-GAN\cite{islam_doa-gan_2020} & 0.4842 & 0.3784 & 0.3692\\
		DFIC\cite{zhong_end--end_2020} & 0.4610 & 0.4220 & 0.4410\\
		Serial Network\cite{chen_serial_2020} & 0.5308 & 0.4979 & 0.4768\\
		Ours & \textbf{0.6296} & \textbf{0.7130} & \textbf{0.6278}\\
		\hline
	\end{tabular}}
\end{table}

\subsection{Comparison to State-of-the-Art Methods}
We compare our method with the state-of-the-art CMFD methods in three datasets: our Synthetic dataset, CASIA CMFD\cite{ferrari_busternet_2018} and CoMoFoD\cite{tralic_comofod_2013}. To be consistent with previous works\cite{ferrari_busternet_2018,islam_doa-gan_2020,chen_serial_2020}, we train our model only on the Synthetic dataset, and the resulting model is used to evaluate the generalization on the other two datasets. Moreover, we only evaluate the localization performance in CASIA CMFD and CoMoFoD datasets since only single-channel masks are offered.

\textbf{Quantitative Analysis:} For the Synthetic dataset, the experimental results are shown in Table \ref{table:Synthetic}, in which $*$ denotes we finetune the model on our Synthetic dataset. Our model achieves the best performance in all three average metrics. Specifically, compared with BusterNet, DOA-GAN, and Serial Network, our method improves the average F1 score by 26\%, 10\%, and 36\%. Moreover, the experimental results in Table \ref{table:casia} and Table \ref{table:comofod} show that the generalization performance of our model is far better than the others. Note that there are many images where copy-move is conducted in background, where previous algorithms overfitted to objects can hardly detect.  To be more specific, our model's F1 metrics outperform the second-best one by over 11\% on CASIA CMFD and over 15\% on CoMoFoD.
\begin{figure}[t!]
	\centering
	\includegraphics[width=1\linewidth]{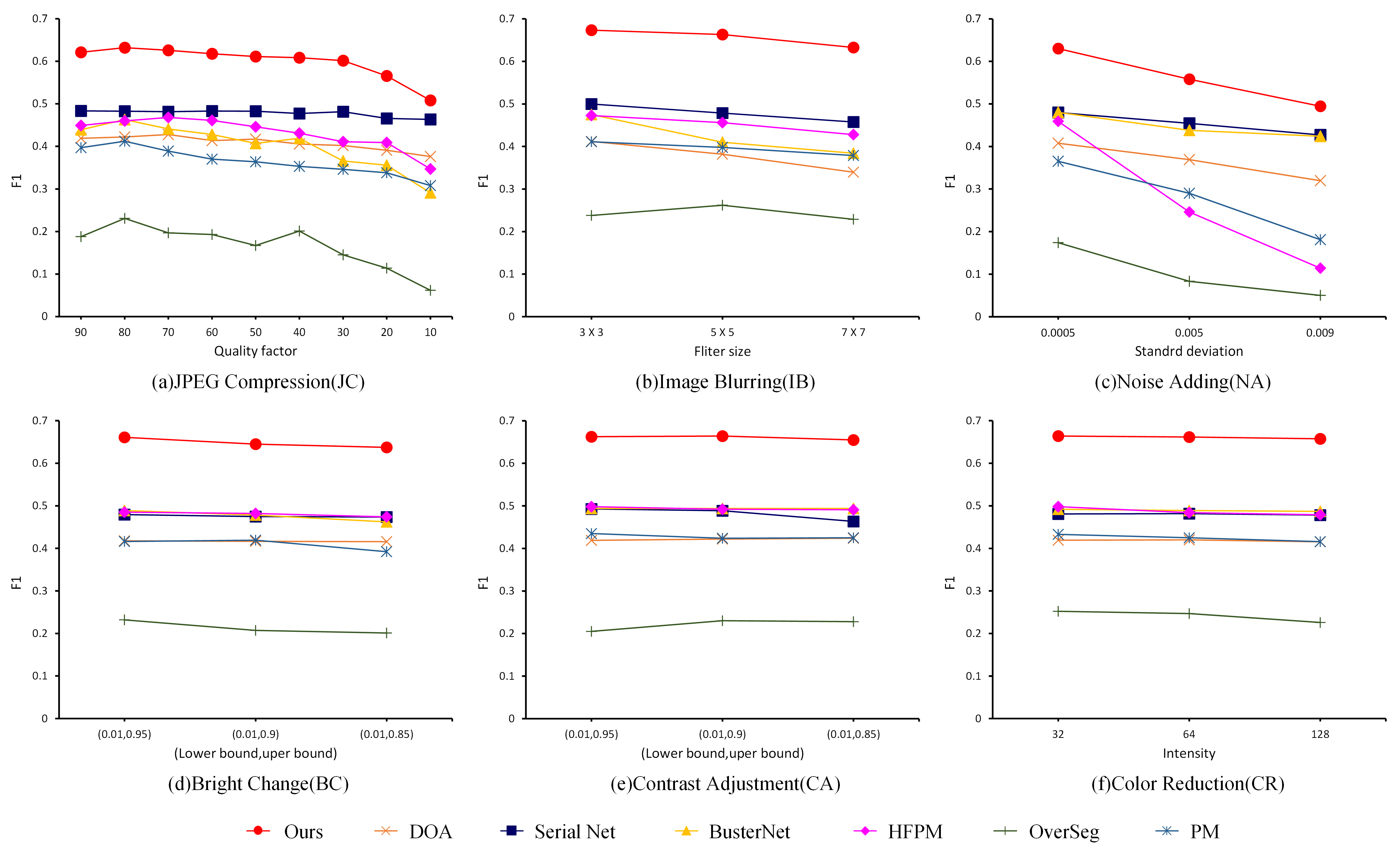}
	\centering
	\caption{F1 comparisons on CoMoFoD dataset under different attacks.}
	\label{fig:comoR}
\end{figure}
\textbf{Qualitative Analysis:} We provide some visual examples in Fig. \ref{fig:visua}.  Fig. 3 (a) and Fig. 3 (b) show that the other three models almost fail to locate the source regions when the copy-move regions belong to the background.  The following Fig. 3 (c) shows the other three models wrongly detect the two people as forged regions, and this further demonstrates they are overfitted to objects.  Compared with those approaches, our model seeks reliable point-to-point matching between source-targert regions, which prevents our method from overfitting to objects. We can see that only our method can successfully locates the copy-move regions when they are not objects. A similar phenomenon can also be seen in Fig. 3 (e).
Fig.3 (d) and Fig. 3 (f) show that, due to we match source and target regions in high-resolution scales, our model can offer a finer mask in a simple case and can locate the small copy-move regions. 

\textbf{Robustness Against Attacks:} To analyze the robustness of our model against different attacks in more detail, Fig. \ref{fig:comoR} depicts the F1 obtained by the algorithms under different attacks. One can see that our model is very robust and achieves consistently much better performance than both the conventional algorithms and deep models. Even though severe JPEG compression and noise addition can destroy the similarity between source and target regions and thus affect the dense matching of source and target regions, our model still achieves better results than previous models.




\subsection{Ablation Study}
To prove the effectiveness of our model structures, we design four variants of our model and the experimental results are summarized in Table \ref{table:ablation} in which CNN means the CNN feature extractor, ZM means the Zernike Moments features extractor, DLF means the dense linear fitting block and Cross-Scale means PM algorithm's cross-scale offset evaluation.

We can see from Table \ref{table:ablation} that both two kinds of features contribute to improving our method's performance. Zernike moments features bring about 10\% improvement and CNN features bring about 7\% improvement on the F1 metric. The reason is Zernike moments offer desirable robustness against rotations which improves CNN's training. Moreover, DLF boosts the F1 by 11\% showing the prior information of the offsets' matching quality is important for CNN decoder to locate the copy-move regions. Finally, we can also see that evaluating the offsets' matching score across features of different scales enhances the location performance by addressing the matching problem between the source regions and the target regions after scaling transformation.

\begin{table}
	\centering
	\caption{Ablation result on CASIA CMFD dataset}
	\label{table:ablation}  
 \scalebox{1}{
	\begin{tabular}{cccc|c}
		\hline
		CNN & ZM & DLF & Cross-Scale & F1 \\ \hline
		 & \checkmark & \checkmark & \checkmark & 0.6809 \\
        \checkmark &  & \checkmark & \checkmark & 0.6512 \\
		\checkmark & \checkmark &  & \checkmark & 0.6433  \\ 
		\checkmark & \checkmark & \checkmark &  & 0.6845 \\ 
        \checkmark & \checkmark & \checkmark & \checkmark & \textbf{0.7542}\\
		\hline
	\end{tabular}}
\end{table}

\section{Conclusion}\label{sec:Conculsion}
In this paper, we propose a novel framework for copy-move forgery detection and localization via deep cross-scale patchmatch. Firstly, we designed a deep cross-scale PM algorithm to help to seek explicit and reliable point-to-point matching between source and target regions using features from high-resolution scales. Moreover, a manipulated regions detection branch is designed for Source/Target Discrimination. Finally, the experimental results demonstrate that our method outperforms state-of-the-art methods.

\bibliographystyle{IEEEbib}
\bibliography{TD}

\end{document}